\documentclass[10pt,twocolumn,letterpaper]{article}

\usepackage{iccv}              

\usepackage{multirow}
\usepackage{pifont}
\usepackage{xcolor,colortbl}
\usepackage{caption}   
\usepackage{float}     
\usepackage[percent]{overpic}  
\usepackage{subcaption}  

\usepackage[accsupp]{axessibility}  

\newlength\savewidth
\newcommand{\tablestyle}[2]{\setlength{\tabcolsep}{#1}\renewcommand{\arraystretch}{#2}\centering\footnotesize}

\newcolumntype{x}[1]{>{\centering\arraybackslash}p{#1pt}}
\newcolumntype{y}[1]{>{\raggedright\arraybackslash}p{#1pt}}
\newcolumntype{z}[1]{>{\raggedleft\arraybackslash}p{#1pt}}

\definecolor{baselinecolor}{gray}{.9}
\newcommand{\baseline}[1]{\cellcolor{baselinecolor}{#1}}

\newcommand{\cmark}{\ding{51}}%
\newcommand{\xmark}{\ding{55}}%

\definecolor{iccvblue}{rgb}{0.21,0.49,0.74}
\usepackage[pagebackref,breaklinks,colorlinks,allcolors=iccvblue]{hyperref}

\def\confName{ICCV}
\def\confYear{2025}

\title{LV-MAE: Learning Long Video Representations through Masked-Embedding Autoencoders}

\author{Ilan Naiman\quad
Emanuel Ben-Baruch\quad 
Oron Anschel\quad
Alon Shoshan\quad \\
Igor Kviatkovsky\quad
Manoj Aggarwal\quad
G\'{e}rard Medioni\\[0.3cm]
Amazon\\
{\tt\small \{naimanil, emanbb, oronans, alonshos, kviat, manojagg, medioni\}@amazon.com}
}

\begin{document}
\maketitle
\definecolor{LightRed}{rgb}{1,0.5,0.5}
\begin{abstract}
In this work, we introduce long-video masked-embedding autoencoders (LV-MAE), a self-supervised learning framework for long video representation.
Our approach treats short- and long-span dependencies as two separate tasks.
Such decoupling allows for a more intuitive video processing where short-span spatiotemporal primitives are first encoded and are then used to capture long-range dependencies across consecutive video segments. 
To achieve this, we leverage advanced off-the-shelf multimodal encoders to extract representations from short segments within the long video, followed by pre-training a masked-embedding autoencoder capturing high-level interactions across segments.
LV-MAE is highly efficient to train and enables the processing of much longer videos by alleviating the constraint on the number of input frames.
Furthermore, unlike existing methods that typically pre-train on short-video datasets, our approach offers self-supervised pre-training using long video samples (\eg, 20+ minutes video clips) at scale.
Using LV-MAE representations, we achieve state-of-the-art results on three long-video benchmarks -- LVU, COIN, and Breakfast -- employing only a simple classification head for either attentive or linear probing.
Finally, to assess LV-MAE pre-training and visualize its reconstruction quality, we leverage the video-language aligned space of short video representations to monitor LV-MAE through video-text retrieval. Code is available at \href{https://github.com/amazon-science/lv-mae}{\textcolor{LightRed}{\texttt{https://github.com/amazon-science/lv-mae}}}.

\end{abstract}    
\vspace{-1.6em}
\section{Introduction}
\label{sec:intro}

In recent years, substantial progress has been achieved in the development of models for short-video representation learning. Numerous  \cite{zhu2024languagebind, Wang2024InternVideo2SV, Chen2023VASTAV, Xue2022CLIPViPAP, Luo2021CLIP4ClipAE, Lin2023VideoLLaVALU, Li2023LLaMAVIDAI, Xu2024PLLaVAP, Yang2023Vid2SeqLP} approaches have been introduced, demonstrating state-of-the-art performance across various tasks such as video classification, action recognition, text-video retrieval, video captioning, and video question answering. 

While models targeting short video clips (5–15 seconds long) are effective at capturing isolated moments, atomic actions, or specific events, they usually struggle to capture long-range temporal dependencies essential for understanding complex, extended narratives. For example, grasping the full emotional journey of characters in a drama or following the progression of a complex plot in a movie often requires viewing the entire film, not just a brief scene.

Meanwhile, developing models that can effectively handle long video content (\eg, 20 minutes long) remains a significant challenge. 
First, most current methods process spatio-temporal tokens at the frame level, restricting the input video length they can handle.
Second, scaling these methods to accommodate longer sequences becomes computationally prohibitive, as training on extended videos demands significant GPU memory, along with increased training cost and time, limiting accessibility for many researchers and organizations.
Third, annotating long videos requires substantial effort, and defining unambiguous annotation guidelines is challenging, often leading to inconsistencies.

\begin{figure*}[t!]
    \centering
    \includegraphics[width=.95\textwidth]{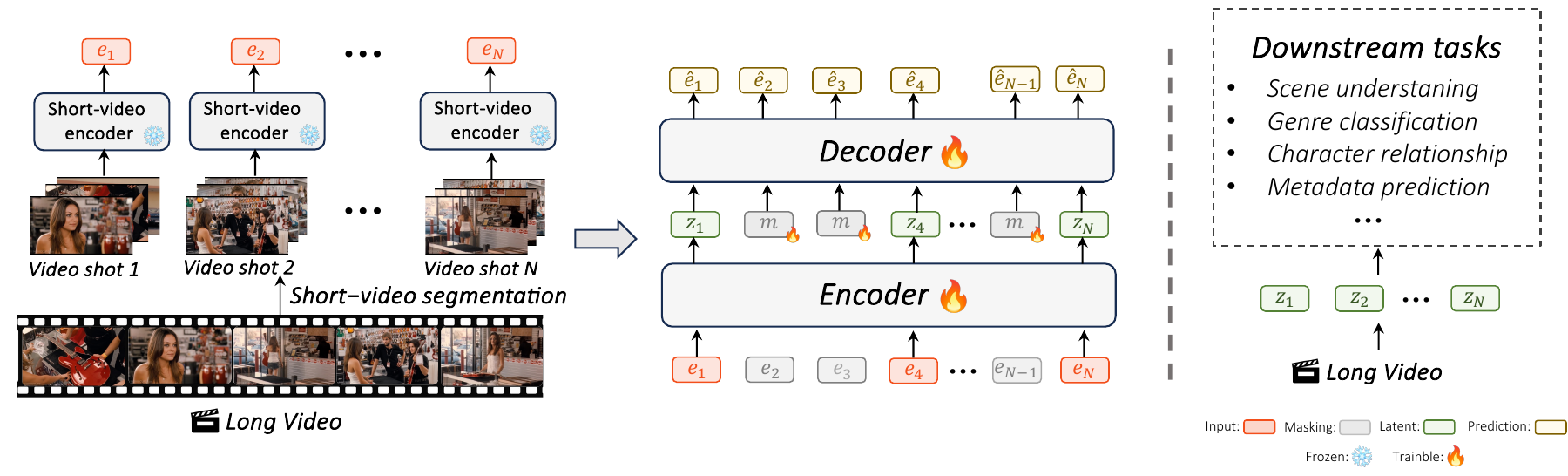}
    \caption{\textbf{Overview of the LV-MAE method.} LV-MAE first utilizes short-video representations extracted by advanced multimodal off-the-shelf encoders (\eg, LanguageBind~\cite{zhu2024languagebind}, InternVideo2~\cite{Wang2024InternVideo2SV}) to capture low-level knowledge of atomic actions and localized events. Next, we pre-train a masked embedding autoencoder to learn long-range dependencies across video segments in a self-supervised manner. After pre-training, the LV-MAE encoder is used to extract high-level representations for long-video downstream tasks.}
    \label{fig:arch}
\end{figure*}

Recently, approaches for handling long-video content have been proposed based on state-space models (SSMs) to address some of the above challenges. ViS4mer~\cite{islam2022long} utilizes a transformer encoder to extract patch-level features from each individual frame, followed by an efficient aggregation of these representations across all frames using the S4 decoder~\cite{gu2022efficiently}. VideoMamba~\cite{li2024videomamba}, a fully SSM-based video model, directly processes spatio-temporal patches from video frames using multiple bidirectional Mamba blocks~\cite{Zhu2024VisionME}. 
These architectures successfully bypass the quadratic complexity of the transformer's attention mechanism and are trained on sequences of up to 64 frames~\cite{islam2022long, li2024videomamba}. Yet, their computational demands rise significantly when applied to videos of extended duration.

In this paper, we tackle the challenges of learning representations for long video content by proposing a self-supervised approach to pre-train on videos ranging from minutes to hours.

Our method is not constrained by the number of frames and is highly efficient in terms of training cost.
Our key idea is to decouple the extraction of low-level representations, which can be effectively achieved using high-performing off-the-shelf short-video models, from the task of modeling long-range dependencies across short video segments.

Specifically, we propose Long-Video MAE (LV-MAE), a transformer-based architecture that uses representations extracted from a sequence of short video segments by off-the-shelf models to learn high-level, long-video representations through self-supervised training.
Our approach begins by segmenting a long video into short clips (\eg, five seconds long) and utilizing a multimodal model (\eg, LanguageBind~\cite{zhu2024languagebind}, IntenVideo2~\cite{Wang2024InternVideo2SV}), pre-trained for video-text alignment, to extract low-level embeddings for each segment. We then train a masked-embedding autoencoder that operates on these low-level embedding tokens to learn a high-level representation of the entire video.
In particular, we adopt the asymmetric encoder-decoder design from~\cite{he2022masked} and demonstrate that reconstructing masked embeddings effectively captures general high-level semantic knowledge across extended video sequences.
Furthermore, we employ a padding strategy and leverage masked self-attention to accommodate videos of arbitrary length.
An overview of the LV-MAE method is provided in~\cref{fig:arch}.

To highlight the difference between our approach and existing methods, consider the tokens required to process a two-minute long video: while frame-level approaches like ViS4mer~\cite{islam2022long} and VideoMamba~\cite{li2024videomamba} require 11,760 tokens (60 frames $\times$ 14 $\times$ 14 patches for a standard 16 $\times$ 16 image grid), our approach processes only 24 tokens in total, with one token per five-second segment.
This significantly reduces the computational burden on the self-attention layer, which has quadratic complexity. 
Furthermore, our method unlocks the capacity to handle a significantly larger number of frames, potentially reaching thousands. Additionally, we bypass the challenge posed by the lack of annotated long-video datasets, as existing datasets contain only short samples of a few minutes. 

A key contribution of our work is enabling pre-training on substantially longer video samples through self-supervised learning. We pre-trained LV-MAE on a large and diverse dataset comprising long-length movies and TV series. Our LV-MAE approach is highly efficient to train, requiring only 2.5 days on a single NVIDIA A10 GPU and 20 hours on 8 NVIDIA A10 GPU.

LV-MAE effectively learns long-range dependencies that generalize well to various downstream tasks. By leveraging the representations learned through LV-MAE, we achieve state-of-the-art performance with simple attentive probing~\cite{yu2022coca} or even linear probing across three long-video benchmarks: LVU~\cite{wu2021towards}, COIN~\cite{tang2019coin}, and Breakfast~\cite{kuehne2014language}. Additionally, we provide interpretable visualizations that monitor the quality of the pre-trained encoder by utilizing the aligned text-video space of the short-video encoder. Specifically, for each reconstructed masked token, we perform retrieval against a large set of captions, resulting in high-quality matches demonstrating the ability to reconstruct the semantic meanings of the masked embeddings. 

Our contributions are summarized as follows: (1) We introduce LV-MAE, a method for learning long-range dependencies across short-video segments by pre-training masked-embedding autoencoders on very long video samples (\eg, 20+ minutes). 

Our method is highly efficient to train and enables the processing of much longer videos by alleviating the constraint on the number of frames.
To the best of our knowledge, this is the first approach to apply masked autoencoders on sequences of embedding vectors instead of image patches.
(2) Leveraging LV-MAE representations, we achieve state-of-the-art results with minimal fine-tuning, using techniques such as attentive or linear probing, on three long-video downstream tasks: LVU~\cite{wu2021towards}, COIN~\cite{tang2019coin}, and Breakfast~\cite{kuehne2014language}. (3) By exploiting the video-language shared space of the short-video encoder, we introduce an interpretable technique to visualize how semantic knowledge is reconstructed within the LV-MAE encoder.
\section{Related Work}
\label{sec:related_work}
\paragraph{Self-supervised learning (SSL).} 
Self-supervised learning (SSL) has brought significant advancements in representation learning by leveraging large-scale unlabeled data~\cite{mikolov2013efficient, radford2018improving, chen2020simple, grill2020bootstrap, caron2021emerging}. In natural language processing (NLP), models like BERT~\cite{devlin2019bert} utilize masked token training strategies to learn contextual embeddings. Inspired by these successes, SSL methods have been adapted to computer vision tasks. In the vision domain, masked image modeling approaches like BEiT~\cite{bao2022beit} and MAE~\cite{he2022masked} have demonstrated remarkable performance in learning visual representations by reconstructing masked image patches. MAE introduces an asymmetric encoder-decoder architecture that reconstructs masked image patches, leading to efficient training and strong downstream performance. In this work, we adapt the MAE asymmetric encoder-decoder architecture and apply it to embeddings instead of images. Similar to how BERT captures semantic context in language by modeling token dependencies, we aim to learn the semantic context between embedding vectors for long-video understanding tasks.
\vspace{-5mm}
\paragraph{Masked autoencoders for video understanding.} VideoMAE~\cite{tong2022videomae} and VideoMAE V2~\cite{wang2023videomae} extend masked autoencoders to video data by reconstructing masked video patches, enabling efficient self-supervised learning. However, they focus on short video clips up to 32 frames, and scaling to longer videos is computationally challenging due to the quadratic complexity of attention mechanisms with respect to sequence length. A decoupling strategy is explored by
CoSeg~\cite{wang2023coseg}, yet it also operates at the \emph{frame} level and therefore, it inherits the same scaling bottleneck. In contrast to these methods, our work addresses long-video understanding by operating on sequences of embeddings and aims to learn long-range dependencies without being constrained by the number of frames. Our novelty lies in the granularity of the low-level representations, leveraging clip-level features to enable efficient training, avoiding costly frame processing.

\vspace{-5mm}
\paragraph{Long-video understanding methods.} Understanding long videos, such as movies or instructional content, is challenging due to the computational cost of processing lengthy sequences and the need to model long-range temporal dependencies. State-space models (SSMs) have been proposed to address some of these challenges. Vis4mer~\cite{islam2022long} combines a transformer encoder for extracting spatial features with a structured state-space sequence model (S4)~\cite{gu2022efficiently} decoder to efficiently aggregate representations across frames. VideoMamba~\cite{li2024videomamba} is a fully SSM-based video model that processes spatiotemporal patches directly using bidirectional Mamba blocks~\cite{Zhu2024VisionME}. While SSMs reduce computational complexity compared to standard transformers, operating directly at the frame level does not fully alleviate computational demands, limiting processing to up to 64 frames~\cite{islam2022long, li2024videomamba}. Our work focuses on enhancing long-video representations through a novel approach, specifically designed for long-form content. We evaluate our LV-MAE method on long-form video classification and regression tasks, demonstrating its effectiveness in handling extended temporal sequences. A detailed discussion of long-video language understanding methods that leverage Large Language Models (LLMs) is provided in App.~\ref{app:related}, as these approaches, while related, address different aspects of the long-video understanding challenge.

\section{Method}
\label{sec:method}

In this section, we introduce the LV-MAE framework for learning long-video representations. Our approach operates hierarchically: first, we extract short-video representations using off-the-shelf models such as LanguageBind~\cite{zhu2024languagebind} and InternVideo2~\cite{Wang2024InternVideo2SV}. Next, we capture long-range dependencies across video segments through self-supervised learning. Specifically, we propose a masked-embedding autoencoder that operates on these short-video embeddings, with an architecture inspired by the asymmetric encoder-decoder design proposed in MAE~\cite{he2022masked}. In the following subsections, we delineate each component of the framework.

\subsection{Short-video Representation} \label{subsec:short_video_representation}
Given a full video $V$, we first segment it into a set of $N$ consecutive short videos (\eg, five seconds long), represented as $\mathcal{V} = \{\mathbf{v}_i\}_{i=1}^{N}$. For each segment, we extract $T$ frames, denoted by $\mathbf{v}_i \in \mathbb{R}^{T \times C \times H \times W}$, where $C$, $H$, and $W$ are the number of channels, height, and width of the frames, respectively. Each video segment $\mathbf{v}_i$ is then processed by a multimodal model, pre-trained for short video-text alignment, such as LanguageBind~\cite{zhu2024languagebind} or InternVideo2 \cite{Wang2024InternVideo2SV}. It produces a set of short-video embeddings $\mathcal{E} = \{\mathbf{e}_i\}_{i=1}^N$, where $\mathbf{e}_i \in \mathbb{R}^{d}$ and $d$ is the embedding dimension.

\subsection{Masked-embedding Autoencoder}

MAE~\cite{he2022masked} and its video-based variants~\cite{tong2022videomae, wang2023videomae} operate directly on video frames, masking a subset of image patches and reconstructing them at the decoder output. In contrast, our approach processes embedding vectors that represent sequential short video segments.

Formally, given a set of short-video embeddings $\mathcal{E}$, we mask a subset $\mathcal{M} \subset \mathcal{E}$, while only the remaining tokens $\mathcal{E} \setminus \mathcal{M}$ are passed as input to the encoder,
\begin{equation} \label{eq:encoding}
    \mathcal{Z} = \text{Encoder}(\mathcal{E} \setminus \mathcal{M}),
\end{equation}
where $\mathcal{Z}$ is the set of encoded visible embeddings. 
We then provide the decoder with both $\mathcal{Z}$ and $|\mathcal{M}|$ mask tokens, where each mask token is represented by a shared, learnable vector, denoted as $\mathbf{m}$. Additionally, we incorporate positional encoding at the encoder and decoder to indicate each token's location within the long video sequence. 

The decoder then outputs a set of reconstructed embeddings $\hat{\mathcal{E}} = \{\hat{\mathbf{e}}_i\}_{i=1}^N$, \ie, 
\begin{equation}
    \hat{\mathcal{E}} = \text{Decoder}\left(\mathcal{Z}, \mathbf{m}, \mathcal{P}\right),
\end{equation}
where $\mathcal{P}=\{\mathbf{p}_i\}_{i=1}^N$ is the set of positional embeddings added to both visible and mask tokens corresponding to their original locations.

\vspace{-4mm}
\paragraph{Loss function.}
We use mean squared error (MSE) loss during training, specifically computing the MSE between the original masked embeddings $\mathbf{e}_i \in \mathcal{M}$ and their corresponding reconstructed embeddings $\hat{\mathbf{e}}_i$ as
\begin{equation}
    \mathcal{L} = \frac{1}{|\mathcal{M}|} \sum_{\mathbf{e}_i \in \mathcal{M}} \left\lVert \mathbf{e}_i - \hat{\mathbf{e}}_i \right\rVert_2^2.
\end{equation}
Although simple, we find that the MSE loss is effective for pre-training long videos represented as sequences of embedding vectors.

\vspace{-4mm}
\paragraph{Masking.}
As previously mentioned, in LV-MAE, we mask a portion (\eg, 50\%) of the sequential embedding tokens by removing a subset $\mathcal{M}$ from the input set of tokens $\mathcal{E}$. In this work, we explore two masking strategies: \textit{random} masking and \textit{semantic} masking. In random masking, the subset $\mathcal{M}$ is simply generated by randomly sampling $|\mathcal{M}|$ embeddings and excluding them from $\mathcal{E}$.

Recently, several works have proposed alternative masking strategies based on masking salient regions within the spatio-temporal tokens~\cite{Li2022SemMAESM, Fan2023MotionGuidedMF, Huang2023MGMAEMG}. Inspired by these efforts, we propose a \textit{semantic} masking strategy that focuses on masking distinct or salient elements within the embedding sequence. Specifically, we employ a simple approach that leverages the semantic information within short-video representations. First, we compute the cosine similarity between each embedding $\mathbf{e}_i$ and its preceding embedding $\mathbf{e}_{i-1}$. Then, we select the embeddings with the lowest cosine similarity values and mask them. This encourages the model to learn more complex dependencies between video segments, enhancing its capacity for in-depth long-video understanding. We demonstrate that in some cases, the semantic strategy yields better results, particularly when employing linear probing for classification.

\vspace{-4mm}
\paragraph{Handling varying video lengths}
To handle videos of varying lengths and to maintain computational efficiency, we adopt techniques from BERT~\cite{devlin2019bert}. Each sequence of short-video segments is limited to a maximum of 256 tokens, with shorter sequences padded using a special \texttt{[PAD]} token. During the self-attention process, we employ an attention mask to prevent these padding tokens from influencing model training. 

\subsection{Training and Data} \label{subsec:train_and_data}
\paragraph{Diverse data sources.} Our model is pre-trained on diverse datasets across multiple video domains, enhancing its robustness and adaptability to various tasks. 
Specifically, we pre-train on over 1,000 long-length movies and TV series, encompassing a wide range of genres, styles, and lengths to ensure comprehensive content diversity.
In addition, we incorporate three publicly available datasets. 
\textit{FineVideo}~\cite{Farré2024FineVideo} includes a vast collection of diverse videos with more than 40,000 videos covering a wide spectrum of categories and domains. With thousands of hours of content, FineVideo represents a rich resource for capturing varied human activities, scenes, and interactions. Additionally, we leverage the train set of \textit{MovieClips}~\cite{wu2021towards} that contains approximately 7,000 video clips, typically ranging from one to three minutes, curated from various movies. It serves as a diverse collection of cinematic content. Finally, \textit{ActivityNet}~\cite{caba2015activitynet}, a dataset that focuses on complex human activities that are relevant to daily life, offering a broad range of action categories and scenarios. Our diverse video dataset supports training across varied domains, enhancing generalization. Since our framework is self-supervised, adding more data is easy and requires no manual labels.

\vspace{-4mm}
\paragraph{Training strategy.} To train our model, we employ a dynamic sampling strategy to handle videos of varying lengths. For each video, we randomly sample a duration in seconds, and this ensures that each batch contains videos of different lengths. It allows us to train on entire videos or portions of them, providing our model with exposure to both short and long video segments, which is particularly useful for handling complex temporal dependencies in long-form video understanding tasks. In practice, for simplicity, we set each short video segment to a length of five seconds. In App.~\ref{app:extended_lvu_ab} we explore the impact of segment length.

\vspace{-4mm}
\paragraph{Training efficiency.} Our pre-training method is highly efficient compared to existing approaches. This efficiency is demonstrated by the reduced number of tokens required during training. Specifically, since we leverage short-video representations extracted by off-the-shelf models, each video segment is represented by a single token. Thus, with each segment lasting five seconds, processing a two-minute video with LV-MAE requires only 24 tokens, compared to the 11,760 tokens required by other frame-level methods (60 frames $\times$ 14 $\times$ 14 patches for a standard 16 $\times$ 16 image grid)~\cite{li2024videomamba, tong2022videomae, wang2023videomae, islam2022long, lin2022learning}.
As a result, we can pre-train LV-MAE in only 20 hours on 8 $\times$ A10 GPUs while achieving state-of-the-art results on downstream tasks. 
Our optimized framework makes pre-training on long videos computationally feasible without compromising performance. Additional implementation details are provided in App.~\ref{app:implementation_details}.

\begin{table*}[h]
    \caption{\textbf{LVU benchmark results.} We compare the Top-1 accuracy results obtained by LV-MAE and other methods. ``FB'' (Frozen Backbone) indicates that only a small subset of parameters is fine-tuned, in contrast to other methods that tune the entire model. ``Dir.'' and ``Rel.'' refer to ``Director'' and ``Relationship,'' respectively. LanguageBind/InternVideo2-Baseline refers to using the raw embeddings.}
    \label{tab:lvu_benchmark}
    \centering
    \small
    \setlength{\tabcolsep}{6.5pt}
    \begin{tabular}[t]{l|c|cccc|ccc|c||cc}
            \hline 
            
            \hline
            
            \hline
        \multirow{2}*{\textbf{Method}} & \multirow{2}*{\shortstack{\textbf{FB}}} & \multicolumn{4}{c|}{\textbf{Metadata $\uparrow$}} & \multicolumn{3}{c|}{\textbf{Content $\uparrow$}} & \multirow{2}*{\textbf{Avg.}} & \multicolumn{2}{c}{\textbf{User $\downarrow$}} \\
        & & \textbf{Dir.} & \textbf{Genre} & \textbf{Writer} & \textbf{Year} & \textbf{Scene} & \textbf{Speak} & \textbf{Rel.} & & \textbf{Likes} & \textbf{Views} \\
        \hline
        VideoBERT~\cite{sun2019videobert} & \xmark & 47.30 & 51.90 & 38.50 & 36.10 & 54.90 & 37.90 & 52.80 & 45.6 & 0.32 & 4.46 \\
        Object Transformer~\cite{wu2021towards} & \xmark & 51.20 & 54.60 & 34.50 & 39.10 & 56.90 & 39.40 & 53.10 & 46.9 & \underline{0.23} & 3.55 \\
        LST~\cite{islam2022long} & \xmark & 56.07 & 52.70 & 42.26 & 39.16 & 62.79 & 37.31 & 52.38 & 48.9 & 0.31 & 3.83 \\
        Performer~\cite{islam2022long} & \xmark & 58.87 & 49.45 & 48.21 & 41.25 & 60.46 & 38.80 & 50.00 & 49.6 & 0.31 & 3.93 \\
        Orthoformer~\cite{islam2022long} & \xmark & 55.14 & 55.79 & 47.02 & 43.35 & 66.27 & 39.30 & 50.00 & 50.9 & 0.29 & 3.86 \\
        ViS4mer~\cite{islam2022long} & \xmark & 62.61 & 54.71 & 48.80 & 44.75 & 67.44 & \underline{40.79} & 57.14 & 53.7 & 0.26 & 3.63 \\
        VideoMamba~\cite{li2024videomamba} & \xmark & \underline{67.29} & 65.24 & 52.98 & 48.23 & 70.37 & 40.43 & \textbf{62.50} & 58.1 & 0.26 & 2.90 \\
        \hline
        LanguageBind-Transformer & \xmark & 24.30 & 57.88 & 16.07 & 18.44 & 35.80 & 32.45 & 51.22 & 33.7 & 0.56 & 3.04 \\
        LanguageBind-Mamba & \xmark & 61.68 & \underline{70.38} & 51.78 & \underline{56.74} & \underline{74.04} & 38.83 & 43.90 & 56.8 & 0.30 & 2.98 \\
        InternVideo2-Baseline~\cite{Wang2024InternVideo2SV} & \cmark & 45.79 & 54.79 & 5.36 & 8.51 & 34.57 & 30.32 & 51.22 & 32.9 & 0.24 & 3.22 \\
        LanguageBind-Baseline~\cite{zhu2024languagebind} & \cmark & 64.48 & 61.47 & 5.95 & 29.08 & 30.86 & 31.91 & 48.78 & 38.9 & \textbf{0.21} & 2.90 \\
        \hline
        \rowcolor{Gray!15} $\text{LV-MAE}_{\text{InternVideo2}}$(Ours) & \cmark & 62.62 &  68.15 & \underline{57.14} & \textbf{58.15} & \textbf{77.78} & 39.89 & 53.66 & \underline{59.6} & \underline{0.23} & \underline{2.68} \\
        \rowcolor{Gray!15} $\text{LV-MAE}_{\text{LanguageBind}}$(Ours) & \cmark & \textbf{77.57} & \textbf{71.57} & \textbf{64.28} & \textbf{58.15} & 72.84 & \textbf{40.95} & \underline{58.53} & \textbf{63.4} & \underline{0.23} & \textbf{2.52} \\
            \hline 
            
            \hline
            
            \hline
    \end{tabular}
    \vskip -0.1in
\end{table*}

\begin{table}[h]
    \caption{\textbf{Breakfast and COIN benchmark results}. We compare the Top-1 accuracy results obtained by LV-MAE and other methods.  ``FB'' (Frozen Backbone) indicates that only a small subset of parameters is fine-tuned, in contrast to other methods that tune the entire model. Baseline refers to using the raw embeddings.}
    \vspace{-1.5em}
    \begin{center}
    \small
        \begin{tabular}{l|c|cc}
            \hline
            
            \hline
            
            \hline\\[-3mm]
            {\multirow{2}{*}{\textbf{Method}}} & {\multirow{2}{*}{\textbf{FB}}} & \textbf{Breakfast} & \textbf{COIN}\\
            & & Top-1 & Top-1 \\
            \hline
            Timeception~\cite{hussein2019timeception} & \xmark & 71.3 & - \\
            VideoGraph~\cite{hussein2019videograph} & \xmark & 69.5 & - \\
            GHRM~\cite{zhou2021graph} & \xmark & 75.5 & - \\
            Distant Supervision~\cite{lin2022learning} & \xmark & 89.9 & 90.0 \\
            ViS4mer~\cite{islam2022long} & \xmark & 88.2 & 88.4 \\
            Turbo~\cite{Han22b} & \xmark & 91.3 & 87.5 \\
            VideoMamba~\cite{li2024videomamba} & \xmark & \textbf{96.9} & 90.4 \\

            \hline
            InternVideo2-Baseline~\cite{Wang2024InternVideo2SV} & \cmark & 83.94 & 88.41 \\
            LanguageBind-Baseline~\cite{zhu2024languagebind} & \cmark & 73.52 & 91.13 \\
            \hline
            \rowcolor{Gray!15} $\text{LV-MAE}_{\text{InternVideo2}}$ (Ours) & \cmark & \underline{93.24} & \textbf{92.72} \\
            \rowcolor{Gray!15} $\text{LV-MAE}_{\text{LanguageBind}}$ (Ours) & \cmark & 91.55 & \underline{92.42} \\
            \hline 
            
            \hline
            
            \hline
        \end{tabular}
    \end{center}
\vspace{-1.2em}
\label{tab:bf_coin_benchmark}
\vspace{-0.5em}
\end{table}

\subsection{Interpretable Predictions} \label{subsec:interpretability}

One potential limitation in training masked-embedding autoencoders, where we process sequences of embeddings rather than spatio-temporal image patches at the frame level, is that it can be challenging to directly interpret the quality of token reconstruction produced by the decoder (\eg as done in \cite{he2022masked}). To address this, we leverage the aligned subspace of language-video embeddings from the pre-trained multimodal model to interpret predictions and assess the model's ability to reconstruct embeddings with accurate semantic meaning.
Specifically, we propose using retrieval for each reconstructed masked token against a large set of captions to visualize and evaluate the matches.
\vspace{-5mm}
\paragraph{Constructing caption database.} We begin by annotating a large set of short video segments from the LVU dataset, utilizing Claude-3.5 Sonnet \cite{anthropic2024claude35sonnet} to automatically generate textual descriptions for each segment. Specifically, we uniformly sample 20 frames from each segment and prompt the LLM to provide a concise description that captures the key visual or semantic information in the segment. Formally, for each segment $v_i$, we obtain an annotation $a_i$.
Next, we use LanguageBind's language encoder to extract textual embedding for each segment's description, $\mathbf{e}_i^{\ell}$.
\vspace{-5mm}
\paragraph{Masked-embeddings retrieval.} Given a reconstructed embeddings output by the decoder $\hat{\mathbf{e}}_j$, corresponding to the $j$-th video segment, we retrieve the top-matching captions with the highest response by computing the cosine similarity between $\hat{\mathbf{e}}_j$ and each caption's textual embedding,
$\mathbf{e}_i^{\ell}$. This way, these top-matched retrieved captions can be inspected to ensure that the semantic meaning of the reconstructed masked embeddings is well captured by the decoder. Visualization examples are provided in section \ref{sec:exp_interpretabitlity}.

\vspace{-0.5em}
\section{Experiments}
\label{sec:experiments}

We present results obtained by our method and compare them to the latest state-of-the-art (SOTA) long video understanding models such as VideoBERT~\cite{sun2019videobert}, ViS4mer~\cite{islam2022long}, Turbo~\cite{Han22b} and VideoMamba~\cite{li2024videomamba}. In particular, we show that LV-MAE outperforms previous SOTA methods on downstream tasks from diverse domains such as movies, activity prediction, and procedural tasks. Additionally, we perform thorough ablation studies to assess the effectiveness of various design choices in our approach. We provide implementation details in App.~\ref{app:implementation_details}.

\subsection{Benchmarks}
\paragraph{Long-Video Understanding (LVU).} A challenging suite of tasks aimed at testing models' capabilities in understanding extended movie clips~\cite{wu2021towards}.
The benchmark consists of nine diverse tasks divided into three main categories: content understanding, metadata prediction, and user engagement prediction.
The \emph{content understanding} tasks involve predicting attributes like ``relationship'', ``speaking style'', and ``scene/place''. \emph{Metadata prediction} tasks focus on identifying movie-related attributes such as ``director'', ``genre'', ``writer'', and ``release year''.
Lastly, the \emph{user engagement} tasks measure how well the model can predict YouTube-based metrics such as the ``like ratio'' and ``popularity''. 
The LVU benchmark is built from the MovieClip dataset that is comprising approximately 30,000 videos sourced from around 3,000 movies. 
Each video spans one to three minutes, reflecting real-world movie scenes that require both short-term and long-term temporal reasoning.
\vspace{-1.5em}
\paragraph{Breakfast.} Video classification of cooking activities.
The dataset~\cite{kuehne2014language} consists of 1,712 videos, covering 10 complex cooking activities and totaling 77 hours of footage.
\vspace{-1.5em}
\paragraph{COmprehensive INstruction video analysis (COIN).}
Video classification of procedural tasks depicted in videos.
The COIN dataset~\cite{tang2019coin} includes 11,827 videos, representing 180 distinct procedural tasks, with an average video length of 2.36 minutes.

\subsection{Main Results}
We evaluate our method on the aforementioned long-video understanding benchmarks, using the top-1 accuracy metric for classification tasks and mean-squared error (MSE) for regression tasks. Initially, we pre-trained our model on a large, diverse video dataset (see Sec.~\ref{subsec:train_and_data} for details). 
To obtain short-video representations, we experimented with either LanguageBind~\cite{zhu2024languagebind} or InternVideo2~\cite{Wang2024InternVideo2SV}, keeping their weights frozen throughout pre-training.
After pre-training, we discard the decoder and use the frozen encoder to extract the long-video representation $\mathcal{Z}$. 
We apply attentive probing (AP)~\cite{yu2022coca, lee2019set} for classification tasks by tuning a single transformer encoder layer with a linear projection head
(details are provided in the appendix), while for regression tasks in LVU, we use a regression model on the global average pooling of the latent representation.

We present our results in Table~\ref{tab:lvu_benchmark} and Table~\ref{tab:bf_coin_benchmark}.
As observed, LV-MAE with LanguageBind embeddings significantly outperforms existing approaches.
Specifically, on the LVU benchmark, our model achieves superior performance in seven out of nine tasks across all categories, with an average accuracy improvement of 5.3\% on the classification tasks. Notably, LV-MAE achieves improvements exceeding 10\% on certain tasks, underscoring the effectiveness of our SSL framework.
On the COIN dataset, LV-MAE outperforms all existing methods using both InternVideo2 and LanguageBind embeddings. 
Finally, on the Breakfast dataset, LV-MAE achieves the second-best performance.

We note that previous works fine-tune all model weights for each benchmark, whereas in our approach, only a small number of parameters are trained through attentive probing. Nevertheless, LV-MAE achieves superior performance on most tasks.
Furthermore, previous works typically pre-train their models on short-video datasets, such as Kinetics~\cite{Kay2017TheKH} (e.g., ViS4mer~\cite{islam2022long} and VideoMamba~\cite{li2024videomamba}). Our approach enables label-free pre-training on long videos by leveraging short-video segments, capturing long-range dependencies and robust temporal patterns.

To further validate the effectiveness of the representations learned by our masked-embedding autoencoder, we compare LV-MAE to a baseline method that applies attentive probing directly to the raw embeddings of either LanguageBind or InternVideo2 (referred to as LanguageBind-Baseline and InternVideo2-Baseline in Table~\ref{tab:lvu_benchmark} and Table~\ref{tab:bf_coin_benchmark}). The results show that while raw embeddings yield reasonable results for certain tasks, our masked-embedding autoencoder consistently produces superior performance, highlighting its effectiveness in capturing high-level context from long videos beyond the raw embedding level.
Notably, on some tasks, using raw InternVideo2 or LanguageBind embeddings leads to poor accuracy (e.g., for the “Writer,” “Year,” and “Scene” tasks). However, pre-training with these embeddings following the LV-MAE scheme significantly improves accuracy. For instance, on the “Writer” task, LV-MAE with LanguageBind improves accuracy from 5.95\% to 64.28\%. Additionally, we conducted end-to-end training experiments using both transformer and Mamba models on LanguageBind embeddings. Results on the LVU benchmark are presented in Table~\ref{tab:lvu_benchmark}. These experiments demonstrate that our self-supervised MAE approach outperforms direct end-to-end training from sequences of short video-clip embeddings.

\subsection{Main Properties} 
We evaluate the influence of various components in our method. Specifically, we investigate the use of linear probing (LP) versus attentive probing (AP) as classification heads, examine the effectiveness of our proposed semantic masking strategy compared to random masking, test different masking ratios, and compare results obtained with various model sizes (encoder depth).
\vspace{-1em}

\paragraph{Attentive probing vs. linear probing.} In Table~\ref{tab:masking_type}, we compare LP and AP using either InternVideo2 or LanguageBind embeddings on the LVU benchmark, Breakfast, and COIN datasets. 
For the LVU benchmark, we report the average score for classification tasks, with full results available in App.~\ref{app:results_downstream_tasks}.
Our findings show that attentive probing consistently outperforms linear probing. This is likely because AP’s additional attention mechanism enables it to capture task-specific context more effectively, whereas LP averages information across the sequence, potentially oversmoothing important details.
For instance, on the Breakfast dataset, attentive probing proves especially useful in distinguishing between cooking activities that share similar steps but differ in subtle nuances.
When using linear probing, averaging the latent representations may smooth out important distinctions between actions, such as cracking eggs directly into a pan versus stirring them first before cooking. 
These minor variations in the long video are critical in tasks like identifying scrambled versus fried eggs, where actions overlap but diverge in essential details. 
Attentive probing preserves these fine-grained nuances across the sequence, making it more effective for such distinctions.

\begin{table}[]
\centering
\caption{\textbf{Masking strategy for linear and attentive probing}. Top-1 accuracy comparison for different masking strategies using InternVideo2 or LanguageBind evaluated on LVU (Avg.), Breakfast (BF), and COIN. Default settings are marked in \colorbox{baselinecolor}{gray}.}
\vspace{-1.5em}
\label{tab:masking_type}
\begin{center}
\tablestyle{10pt}{1.05}
\begin{tabular}{llccc}
\hline 

\hline

\hline
\textbf{Masking} & \textbf{Embedding} & \textbf{LVU} & \textbf{BF} & \textbf{COIN} \\
\hline
\multicolumn{5}{c}{Linear Probing} \\
\hline
 Semantic & LanguageBind & \textbf{59.02} & 81.97 & 91.54 \\ 
 Random & LanguageBind & 58.32 & 81.69 & 91.46 \\
\hline
Semantic & InternVideo2 & 52.60 & \textbf{85.92} & \textbf{91.63} \\ 
Random & InternVideo2 & 53.63 & 85.63 & 91.29 \\
\hline
\multicolumn{5}{c}{Attentive Probing} \\
\hline
Semantic & LanguageBind & \baseline{\textbf{63.41}} & \baseline{90.70} & \baseline{91.8} \\
Random & LanguageBind &  62.54 & 91.55 & 92.42 \\
\hline
Semantic & InternVideo2 & 59.63 & \textbf{93.24} & 92.72 \\
Random & InternVideo2 & 58.43  & 92.68 & \textbf{92.88} \\

\hline 

\hline

\hline
\end{tabular}
\end{center}
\vspace{-2.5em}
\end{table}

\begin{figure}[]
    \centering
    \includegraphics[width=.45 \textwidth]{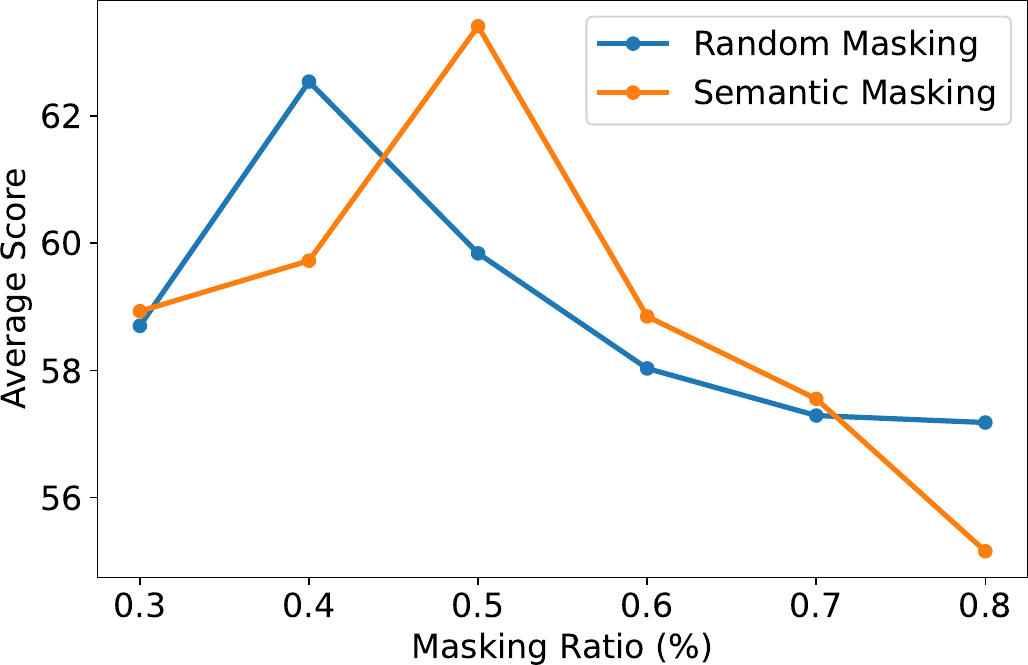}
    \caption{\textbf{Masking ratio}. The y-axes are LVU average accuracy scores. A moderate masking ratio of $40-50\%$ works well for attentive probing.}
    \label{fig:cos_mask_ablation}
    \vspace{-1.0em}
\end{figure}

\vspace{-1.5em}
\paragraph{Masking strategy.}  We compare our proposed semantic masking strategy against the random masking strategy as employed by the standrad MAE model. 
As shown in Table~\ref{tab:masking_type}, semantic masking consistently performs slightly better than random masking, suggesting that selectively masking more semantically relevant embeddings enhances model performance.
We observe that both masking strategies are simple yet effective, achieving state-of-the-art results. 
\vspace{-1.5em}
\paragraph{Masking ratio.}
Additionally, we explore various masking ratios and find that moderate masking ratios yield the best results, with $40\%$ for random masking and $50\%$ for semantic masking producing the highest performance (Figure~\ref{fig:cos_mask_ablation}). 
Finally, we assess the effect of model size on performance.
\vspace{-1.5em}
\paragraph{Model size.} 
Table~\ref{tab:encoder_depth} reports accuracy results for different numbers of transformer encoder blocks. As shown, increasing the number of encoder blocks improves accuracy in nearly all tested cases, indicating that deeper encoders lead to stronger performance.

\begin{table}[]
\centering
\caption{\textbf{Encoder depth}. A deep encoder can improve accuracy. Here, we use attentive probing and Top 1 accuracy. We evaluated on LVU (Avg.), Breakfast (BF), and COIN datasets. Default settings are marked in \colorbox{baselinecolor}{gray}.}
\vspace{-1.5em}
\begin{center}
\label{tab:encoder_depth}
\tablestyle{10pt}{1.05}
\begin{tabular}{cccc}
\hline

\hline

\hline
\textbf{Depth} & \textbf{LVU} & \textbf{BF} & \textbf{COIN} \\
\hline
16 & 59.73 & 87.04 & 92.47 \\
24 & 60.04 & 89.58 & \textbf{93.26} \\
32 & \baseline{\textbf{63.41}} & \baseline{\textbf{90.70}} & \baseline{91.8} \\
\hline

\hline

\hline
\end{tabular}
\end{center}
\vspace{-2.0em}
\end{table}

\begin{figure*}[ht]
    \centering
    \includegraphics[width=1.\textwidth]{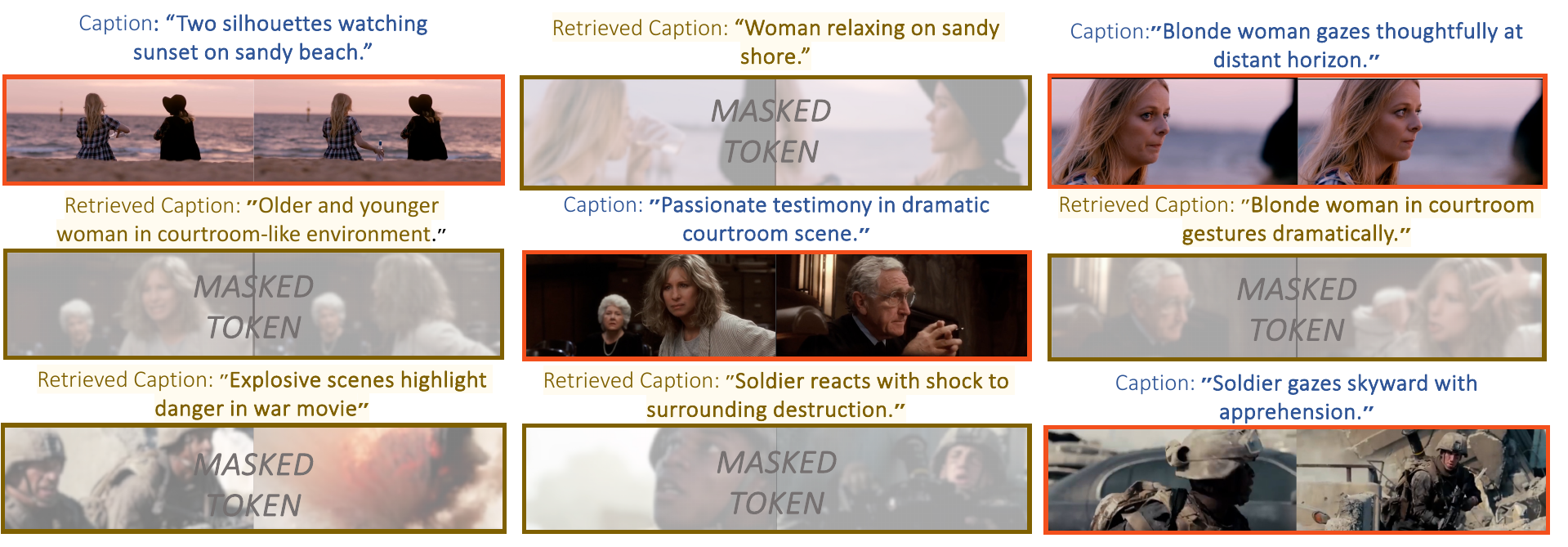} 
    \caption{\textbf{Interpretable predictions -- examples}: Each row visualizes three consecutive five-second segments. Above each segment, we show the original caption for the visible tokens and the retrieved caption for the reconstructed masked tokens. As shown, the model successfully reconstructs the semantic meaning of the masked embeddings, offering insight into the model's effectiveness and capabilities.}
    \label{fig:interp_res}
\end{figure*}

\begin{figure}[ht]
    \hspace{1cm}
    \includegraphics[width=.5\textwidth]{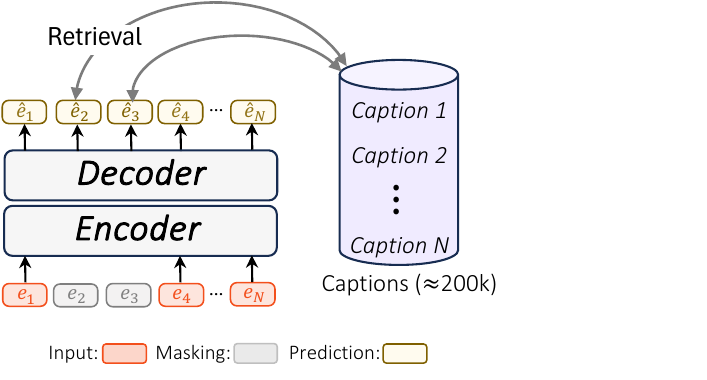}
    \caption{\textbf{Interpretable predictions -- scheme}: For each reconstructed masked embedding, we perform retrieval against a large set of captions collected from the MovieClip dataset. The top matches can then be used to assess the model's quality.}
    \label{fig:interp_scheme}
\vspace{-2.0em}
\end{figure}

\subsection{Interpretable Predictions Experiment}
\label{sec:exp_interpretabitlity}
The ability to understand model predictions is essential to facilitate human understanding of what the model learns. In Sec.~\ref{subsec:interpretability} we propose an interpretability technique for our framework. In this section, we show its effectiveness and verify the reconstructions produced by the decoder. 
In practice, we used the MovieClips dataset to extract captions. For each short-video segment, $v_i$, in the dataset, we generate concise annotations, $a_i$, using the Claude Sonnet 3.5 LLM \cite{anthropic2024claude35sonnet}. Next, we sample a long video from the dataset and extract its corresponding embeddings, $\mathcal{E}$, masking $40\%$ of them. Our model is fed with the visible embeddings and tasked with reconstructing all embeddings, resulting in the predicted set, $\hat{\mathcal{E}}$. Each annotation is transformed into a language embedding using LanguageBind, which has the same embedding space as $\mathcal{E}$. Using the cosine similarity, we perform retrieval by selecting the most similar textual embeddings. We conduct this retrieval for both the original embeddings, $\mathcal{E}$, and the predicted embeddings, $\hat{\mathcal{E}}$. We provide a scheme of the process in Figure~\ref{fig:interp_scheme}. The original model retrieval accuracy (R@5) is 75.34\%.

For concise visualization of the results, in Fig.~\ref{fig:interp_res}, we subsample short video segments and present the interpreted model's predictions. We observe that in most cases, our model predicts embeddings that are semantically close to the original ones. This means that the predicted embeddings align with either the same or semantically similar enough textual embedding. As expected, we observe that the retrieved textual embeddings tend to be more abstract or high-level, rather than precise descriptions of the scene. For instance, in one example, the ground truth annotation is ``Blonde woman sips drink by the sea'', while the retrieved annotation according to the prediction is "Woman relaxing on sandy shore." This behavior is akin to what has been observed with MAE models in the image domain~\cite{he2022masked}, where predicted image patches are often blurry and lack fine details. Similarly, our model’s predictions can be seen as abstract representations, capturing the essence of a scene but missing some specific details. 

\vspace{-0.8em}
\section{Discussion and Limitations}
\label{sec:disc_lim}
\vspace{-0.5em}

Our method achieves state-of-the-art results across diverse tasks and benchmarks, yet there are several limitations to consider. 
First, our approach relies on leveraging low-level representations from pre-trained, off-the-shelf models. 
Since these representations are frozen during our training, our model's performance is inherently constrained by the quality of these pre-trained embeddings. 
If the pre-trained model fails to effectively capture the content of short video clips, this deficiency may propagate into our long-range understanding framework, affecting overall representation quality. 
However, by choosing high-performing video-text alignment models such as LanguageBind, we mitigate this limitation and ensure strong baseline representations.

Second, unlike methods operating on image patches, where masked token predictions can be easily visualized, our approach operates on embeddings, which are inherently more abstract and less interpretable.
This makes it challenging to directly assess the model's performance on masked token predictions. 
To address this, we propose an interpretability strategy that leverages captions or generates them when unavailable, providing a proxy for visualizing and understanding the model’s predictions. 
This approach aids in interpreting predictions, though it remains an approximation rather than a direct visualization of learned embeddings.
\vspace{-0.4em}
\section{Conclusions}
\label{sec:conclusions}
\vspace{-0.4em}

While new video understanding methods are emerging rapidly, most are confined to short clips and operate at the frame level, limiting their scalability to longer content. In this work, we introduce a novel self-supervised representation learning model for long-video understanding. Our method builds on the well-established masked autoencoder design, utilizes high-quality embeddings from pre-trained short-video models, and efficiently handles videos ranging from a few seconds to potentially several hours. This allows us to capture richer temporal dynamics and semantic structures over extended durations. Empirical results show that our model consistently outperforms or matches state-of-the-art approaches on long-video tasks. Looking ahead, our framework offers strong potential for broader applications such as generation, retrieval, and comprehensive long-form analysis.

\clearpage
{
    \small
    \bibliographystyle{ieeenat_fullname}
    \bibliography{main}

\begin{thebibliography}{56}
\providecommand{\natexlab}[1]{#1}
\providecommand{\url}[1]{\texttt{#1}}
\expandafter\ifx\csname urlstyle\endcsname\relax
  \providecommand{\doi}[1]{doi: #1}\else
  \providecommand{\doi}{doi: \begingroup \urlstyle{rm}\Url}\fi

\bibitem[Anthropic(2024)]{anthropic2024claude35sonnet}
Anthropic.
\newblock Introducing claude 3.5 sonnet.
\newblock \emph{Anthropic News}, 2024.

\bibitem[Arnab et~al.(2021)Arnab, Dehghani, Heigold, Sun, Lu{\v{c}}i{\'c}, and Schmid]{arnab2021vivit}
Anurag Arnab, Mostafa Dehghani, Georg Heigold, Chen Sun, Mario Lu{\v{c}}i{\'c}, and Cordelia Schmid.
\newblock Vivit: A video vision transformer.
\newblock In \emph{Proceedings of the IEEE/CVF international conference on computer vision}, pages 6836--6846, 2021.

\bibitem[Bao et~al.(2022)Bao, Dong, Piao, and Wei]{bao2022beit}
Hangbo Bao, Li Dong, Songhao Piao, and Furu Wei.
\newblock {BE}it: {BERT} pre-training of image transformers.
\newblock In \emph{International Conference on Learning Representations}, 2022.

\bibitem[Caron et~al.(2021)Caron, Touvron, Misra, J{\'e}gou, Mairal, Bojanowski, and Joulin]{caron2021emerging}
Mathilde Caron, Hugo Touvron, Ishan Misra, Herv{\'e} J{\'e}gou, Julien Mairal, Piotr Bojanowski, and Armand Joulin.
\newblock Emerging properties in self-supervised vision transformers.
\newblock In \emph{Proceedings of the IEEE/CVF international conference on computer vision}, pages 9650--9660, 2021.

\bibitem[Chen et~al.(2023)Chen, Li, Wang, Zhao, Sun, Zhu, and Liu]{Chen2023VASTAV}
Sihan Chen, Handong Li, Qunbo Wang, Zijia Zhao, Ming-Ting Sun, Xinxin Zhu, and J. Liu.
\newblock Vast: A vision-audio-subtitle-text omni-modality foundation model and dataset.
\newblock \emph{ArXiv}, abs/2305.18500, 2023.

\bibitem[Chen et~al.(2020)Chen, Kornblith, Norouzi, and Hinton]{chen2020simple}
Ting Chen, Simon Kornblith, Mohammad Norouzi, and Geoffrey Hinton.
\newblock A simple framework for contrastive learning of visual representations.
\newblock In \emph{Proceedings of the 37th International Conference on Machine Learning (ICML)}, pages 1597--1607. PMLR, 2020.

\bibitem[Chen et~al.(2024)Chen, Xue, Li, Hu, Zhu, Li, Fang, Tang, Yang, Liu, et~al.]{chen2024longvila}
Yukang Chen, Fuzhao Xue, Dacheng Li, Qinghao Hu, Ligeng Zhu, Xiuyu Li, Yunhao Fang, Haotian Tang, Shang Yang, Zhijian Liu, et~al.
\newblock Longvila: Scaling long-context visual language models for long videos.
\newblock \emph{arXiv preprint arXiv:2408.10188}, 2024.

\bibitem[Devlin et~al.(2019)Devlin, Chang, Lee, and Toutanova]{devlin2019bert}
Jacob Devlin, Ming-Wei Chang, Kenton Lee, and Kristina Toutanova.
\newblock {BERT}: Pre-training of deep bidirectional transformers for language understanding.
\newblock In \emph{Proceedings of the 2019 Conference of the North {A}merican Chapter of the Association for Computational Linguistics: Human Language Technologies, Volume 1 (Long and Short Papers)}, pages 4171--4186, Minneapolis, Minnesota, 2019. Association for Computational Linguistics.

\bibitem[Fabian Caba~Heilbron and Niebles(2015)]{caba2015activitynet}
Bernard~Ghanem Fabian Caba~Heilbron, Victor~Escorcia and Juan~Carlos Niebles.
\newblock Activitynet: A large-scale video benchmark for human activity understanding.
\newblock In \emph{Proceedings of the IEEE Conference on Computer Vision and Pattern Recognition}, pages 961--970, 2015.

\bibitem[Fan et~al.(2023)Fan, Wang, Liao, Zhu, Bhat, Santos-Villalobos, Rohith, and Li]{Fan2023MotionGuidedMF}
David Fan, Jue Wang, Shuai Liao, Yi Zhu, Vimal Bhat, Hector~J. Santos-Villalobos, M.~V. Rohith, and Xinyu Li.
\newblock Motion-guided masking for spatiotemporal representation learning.
\newblock \emph{2023 IEEE/CVF International Conference on Computer Vision (ICCV)}, pages 5596--5606, 2023.

\bibitem[Fan et~al.(2021)Fan, Xiong, Mangalam, Li, Yan, Malik, and Feichtenhofer]{fan2021multiscale}
Haoqi Fan, Bo Xiong, Karttikeya Mangalam, Yanghao Li, Zhicheng Yan, Jitendra Malik, and Christoph Feichtenhofer.
\newblock Multiscale vision transformers.
\newblock In \emph{Proceedings of the IEEE/CVF international conference on computer vision}, pages 6824--6835, 2021.

\bibitem[Farré et~al.(2024)Farré, Marafioti, Tunstall, Von~Werra, and Wolf]{Farré2024FineVideo}
Miquel Farré, Andi Marafioti, Lewis Tunstall, Leandro Von~Werra, and Thomas Wolf.
\newblock Finevideo.
\newblock \url{https://huggingface.co/datasets/HuggingFaceFV/finevideo}, 2024.

\bibitem[Grill et~al.(2020)Grill, Strub, Altch{\'e}, Tallec, Richemond, Buchatskaya, Doersch, Avila~Pires, Guo, Gheshlaghi~Azar, Piot, Cavalier, Gelly, Munos, and Valko]{grill2020bootstrap}
Jean-Bastien Grill, Florian Strub, Florent Altch{\'e}, Corentin Tallec, Pierre Richemond, Elena Buchatskaya, Carl Doersch, Bernardo Avila~Pires, Zhaohan Guo, Mohammad Gheshlaghi~Azar, Bilal Piot, Steffen Cavalier, Sylvain Gelly, Remi Munos, and Michal Valko.
\newblock Bootstrap your own latent: A new approach to self-supervised learning.
\newblock \emph{Advances in Neural Information Processing Systems}, 33:\penalty0 21271--21284, 2020.

\bibitem[Gu et~al.(2022)Gu, Goel, and Re]{gu2022efficiently}
Albert Gu, Karan Goel, and Christopher Re.
\newblock Efficiently modeling long sequences with structured state spaces.
\newblock In \emph{International Conference on Learning Representations}, 2022.

\bibitem[Han et~al.(2022)Han, Xie, and Zisserman]{Han22b}
Tengda Han, Weidi Xie, and Andrew Zisserman.
\newblock Turbo training with token dropout.
\newblock In \emph{British Machine Vision Conference}, 2022.

\bibitem[He et~al.(2022)He, Chen, Xie, Li, Doll{\'a}r, and Girshick]{he2022masked}
Kaiming He, Xinlei Chen, Saining Xie, Yanghao Li, Piotr Doll{\'a}r, and Ross Girshick.
\newblock Masked autoencoders are scalable vision learners.
\newblock In \emph{Proceedings of the IEEE/CVF conference on computer vision and pattern recognition}, pages 16000--16009, 2022.

\bibitem[Huang et~al.(2023)Huang, Zhao, Zhang, Qiao, and Wang]{Huang2023MGMAEMG}
Bingkun Huang, Zhiyu Zhao, Guozhen Zhang, Y. Qiao, and Limin Wang.
\newblock Mgmae: Motion guided masking for video masked autoencoding.
\newblock \emph{2023 IEEE/CVF International Conference on Computer Vision (ICCV)}, pages 13447--13458, 2023.

\bibitem[Huang et~al.(2024)Huang, Liao, Radhakrishnan, Yin, Molchanov, Yu, and Kautz]{huang2024lita}
De-An Huang, Shijia Liao, Subhashree Radhakrishnan, Hongxu Yin, Pavlo Molchanov, Zhiding Yu, and Jan Kautz.
\newblock Lita: Language instructed temporal-localization assistant.
\newblock In \emph{European Conference on Computer Vision}, pages 202--218. Springer, 2024.

\bibitem[Hussein et~al.(2019{\natexlab{a}})Hussein, Gavves, and Smeulders]{hussein2019timeception}
Noureldien Hussein, Efstratios Gavves, and Arnold~WM Smeulders.
\newblock Timeception for complex action recognition.
\newblock In \emph{Proceedings of the IEEE/CVF Conference on Computer Vision and Pattern Recognition}, pages 254--263, 2019{\natexlab{a}}.

\bibitem[Hussein et~al.(2019{\natexlab{b}})Hussein, Gavves, and Smeulders]{hussein2019videograph}
Noureldien Hussein, Efstratios Gavves, and Arnold~WM Smeulders.
\newblock Videograph: Recognizing minutes-long human activities in videos.
\newblock \emph{arXiv preprint arXiv:1905.05143}, 2019{\natexlab{b}}.

\bibitem[Islam and Bertasius(2022)]{islam2022long}
Md~Mohaiminul Islam and Gedas Bertasius.
\newblock Long movie clip classification with state-space video models.
\newblock In \emph{European Conference on Computer Vision}, pages 87--104. Springer, 2022.

\bibitem[Kay et~al.(2017)Kay, Carreira, Simonyan, Zhang, Hillier, Vijayanarasimhan, Viola, Green, Back, Natsev, Suleyman, and Zisserman]{Kay2017TheKH}
Will Kay, Jo{\~a}o Carreira, Karen Simonyan, Brian Zhang, Chloe Hillier, Sudheendra Vijayanarasimhan, Fabio Viola, Tim Green, Trevor Back, Apostol Natsev, Mustafa Suleyman, and Andrew Zisserman.
\newblock The kinetics human action video dataset.
\newblock \emph{ArXiv}, abs/1705.06950, 2017.

\bibitem[Kuehne et~al.(2014)Kuehne, Arslan, and Serre]{kuehne2014language}
Hilde Kuehne, Ali Arslan, and Thomas Serre.
\newblock The language of actions: Recovering the syntax and semantics of goal-directed human activities.
\newblock In \emph{Proceedings of the IEEE conference on computer vision and pattern recognition}, pages 780--787, 2014.

\bibitem[Lee et~al.(2019)Lee, Lee, Kim, Kosiorek, Choi, and Teh]{lee2019set}
Juho Lee, Yoonho Lee, Jungtaek Kim, Adam Kosiorek, Seungjin Choi, and Yee~Whye Teh.
\newblock Set transformer: A framework for attention-based permutation-invariant neural networks.
\newblock In \emph{International conference on machine learning}, pages 3744--3753. PMLR, 2019.

\bibitem[Lei et~al.(2021)Lei, Li, Zhou, Gan, Berg, Bansal, and Liu]{lei2021less}
Jie Lei, Linjie Li, Luowei Zhou, Zhe Gan, Tamara~L Berg, Mohit Bansal, and Jingjing Liu.
\newblock Less is more: Clipbert for video-and-language learning via sparse sampling.
\newblock In \emph{Proceedings of the IEEE/CVF conference on computer vision and pattern recognition}, pages 7331--7341, 2021.

\bibitem[Li et~al.(2022)Li, Zheng, Liu, Su, and Zheng]{Li2022SemMAESM}
Gang Li, Heliang Zheng, Daqing Liu, Bing Su, and Changwen Zheng.
\newblock Semmae: Semantic-guided masking for learning masked autoencoders.
\newblock \emph{ArXiv}, abs/2206.10207, 2022.

\bibitem[Li et~al.(2024{\natexlab{a}})Li, Li, Wang, He, Wang, Wang, and Qiao]{li2024videomamba}
Kunchang Li, Xinhao Li, Yi Wang, Yinan He, Yali Wang, Limin Wang, and Yu Qiao.
\newblock Videomamba: State space model for efficient video understanding, 2024{\natexlab{a}}.

\bibitem[Li et~al.(2023)Li, Wang, and Jia]{Li2023LLaMAVIDAI}
Yanwei Li, Chengyao Wang, and Jiaya Jia.
\newblock Llama-vid: An image is worth 2 tokens in large language models.
\newblock In \emph{European Conference on Computer Vision}, 2023.

\bibitem[Li et~al.(2024{\natexlab{b}})Li, Wang, and Jia]{li2024llama}
Yanwei Li, Chengyao Wang, and Jiaya Jia.
\newblock Llama-vid: An image is worth 2 tokens in large language models.
\newblock In \emph{European Conference on Computer Vision}, pages 323--340. Springer, 2024{\natexlab{b}}.

\bibitem[Lin et~al.(2023)Lin, Zhu, Ye, Ning, Jin, and Yuan]{Lin2023VideoLLaVALU}
Bin Lin, Bin Zhu, Yang Ye, Munan Ning, Peng Jin, and Li Yuan.
\newblock Video-llava: Learning united visual representation by alignment before projection.
\newblock \emph{ArXiv}, abs/2311.10122, 2023.

\bibitem[Lin et~al.(2022)Lin, Petroni, Bertasius, Rohrbach, Chang, and Torresani]{lin2022learning}
Xudong Lin, Fabio Petroni, Gedas Bertasius, Marcus Rohrbach, Shih-Fu Chang, and Lorenzo Torresani.
\newblock Learning to recognize procedural activities with distant supervision.
\newblock In \emph{Proceedings of the IEEE/CVF Conference on Computer Vision and Pattern Recognition}, pages 13853--13863, 2022.

\bibitem[Luo et~al.(2021{\natexlab{a}})Luo, Ji, Zhong, Chen, Lei, Duan, and Li]{Luo2021CLIP4ClipAE}
Huaishao Luo, Lei Ji, Ming Zhong, Yang Chen, Wen Lei, Nan Duan, and Tianrui Li.
\newblock Clip4clip: An empirical study of clip for end to end video clip retrieval.
\newblock \emph{Neurocomputing}, 508:\penalty0 293--304, 2021{\natexlab{a}}.

\bibitem[Luo et~al.(2021{\natexlab{b}})Luo, Ji, Zhong, Chen, Lei, Duan, and Li]{luo2021clip4clip}
Huaishao Luo, Lei Ji, Ming Zhong, Yang Chen, Wen Lei, Nan Duan, and Tianrui Li.
\newblock Clip4clip: An empirical study of clip for end to end video clip retrieval.
\newblock \emph{arXiv preprint arXiv:2104.08860}, 2021{\natexlab{b}}.

\bibitem[Mikolov et~al.(2013)Mikolov, Chen, Corrado, and Dean]{mikolov2013efficient}
Tomas Mikolov, Kai Chen, Greg Corrado, and Jeffrey Dean.
\newblock Efficient estimation of word representations in vector space.
\newblock \emph{arXiv preprint arXiv:1301.3781}, 2013.
\newblock Available at: \url{https://arxiv.org/abs/1301.3781}.

\bibitem[Radford et~al.(2018)Radford, Narasimhan, Salimans, and Sutskever]{radford2018improving}
Alec Radford, Karthik Narasimhan, Tim Salimans, and Ilya Sutskever.
\newblock Improving language understanding by generative pre-training.
\newblock \emph{OpenAI preprint}, 2018.

\bibitem[Radford et~al.(2021)Radford, Kim, Hallacy, Ramesh, Goh, Agarwal, Sastry, Askell, Mishkin, Clark, et~al.]{radford2021learning}
Alec Radford, Jong~Wook Kim, Chris Hallacy, Aditya Ramesh, Gabriel Goh, Sandhini Agarwal, Girish Sastry, Amanda Askell, Pamela Mishkin, Jack Clark, et~al.
\newblock Learning transferable visual models from natural language supervision.
\newblock In \emph{International conference on machine learning}, pages 8748--8763. PMLR, 2021.

\bibitem[Ren et~al.(2024)Ren, Yao, Li, Sun, and Hou]{ren2024timechat}
Shuhuai Ren, Linli Yao, Shicheng Li, Xu Sun, and Lu Hou.
\newblock Timechat: A time-sensitive multimodal large language model for long video understanding.
\newblock In \emph{Proceedings of the IEEE/CVF Conference on Computer Vision and Pattern Recognition}, pages 14313--14323, 2024.

\bibitem[Shu et~al.(2024)Shu, Zhang, Liu, Qin, Zhou, Huang, and Zhao]{shu2024video}
Yan Shu, Peitian Zhang, Zheng Liu, Minghao Qin, Junjie Zhou, Tiejun Huang, and Bo Zhao.
\newblock Video-xl: Extra-long vision language model for hour-scale video understanding.
\newblock \emph{arXiv preprint arXiv:2409.14485}, 2024.

\bibitem[Sun et~al.(2019)Sun, Myers, Vondrick, Murphy, and Schmid]{sun2019videobert}
Chen Sun, Austin Myers, Carl Vondrick, Kevin Murphy, and Cordelia Schmid.
\newblock Videobert: A joint model for video and language representation learning.
\newblock In \emph{Proceedings of the IEEE/CVF international conference on computer vision}, pages 7464--7473, 2019.

\bibitem[Tang et~al.(2019)Tang, Ding, Rao, Zheng, Zhang, Zhao, Lu, and Zhou]{tang2019coin}
Yansong Tang, Dajun Ding, Yongming Rao, Yu Zheng, Danyang Zhang, Lili Zhao, Jiwen Lu, and Jie Zhou.
\newblock {COIN}: A large-scale dataset for comprehensive instructional video analysis.
\newblock In \emph{Proceedings of the IEEE/CVF Conference on Computer Vision and Pattern Recognition}, pages 1207--1216, 2019.

\bibitem[Tong et~al.(2022)Tong, Song, Wang, and Wang]{tong2022videomae}
Zhan Tong, Yibing Song, Jue Wang, and Limin Wang.
\newblock Videomae: Masked autoencoders are data-efficient learners for self-supervised video pre-training.
\newblock \emph{Advances in neural information processing systems}, 35:\penalty0 10078--10093, 2022.

\bibitem[Wang et~al.(2023{\natexlab{a}})Wang, Huang, Zhao, Tong, He, Wang, Wang, and Qiao]{wang2023videomae}
Limin Wang, Bingkun Huang, Zhiyu Zhao, Zhan Tong, Yinan He, Yi Wang, Yali Wang, and Yu Qiao.
\newblock Videomae v2: Scaling video masked autoencoders with dual masking.
\newblock In \emph{Proceedings of the IEEE/CVF Conference on Computer Vision and Pattern Recognition}, pages 14549--14560, 2023{\natexlab{a}}.

\bibitem[Wang et~al.(2023{\natexlab{b}})Wang, Liu, Mei, and Luo]{wang2023coseg}
Xiao Wang, Jingen Liu, Tao Mei, and Jiebo Luo.
\newblock Coseg: Cognitively inspired unsupervised generic event segmentation.
\newblock \emph{IEEE Transactions on Neural Networks and Learning Systems}, 35\penalty0 (9):\penalty0 12507--12517, 2023{\natexlab{b}}.

\bibitem[Wang et~al.(2024)Wang, Li, Li, Yu, He, Chen, Pei, Zheng, Xu, Wang, Shi, Jiang, Li, Zhang, Huang, Qiao, Wang, and Wang]{Wang2024InternVideo2SV}
Yi Wang, Kunchang Li, Xinhao Li, Jiashuo Yu, Yinan He, Guo Chen, Baoqi Pei, Rongkun Zheng, Jilan Xu, Zun Wang, Yansong Shi, Tianxiang Jiang, Songze Li, Hongjie Zhang, Yifei Huang, Yu Qiao, Yali Wang, and Limin Wang.
\newblock Internvideo2: Scaling video foundation models for multimodal video understanding.
\newblock \emph{ArXiv}, abs/2403.15377, 2024.

\bibitem[Weng et~al.(2024)Weng, Han, He, Chang, and Zhuang]{weng2024longvlm}
Yuetian Weng, Mingfei Han, Haoyu He, Xiaojun Chang, and Bohan Zhuang.
\newblock Longvlm: Efficient long video understanding via large language models.
\newblock In \emph{European Conference on Computer Vision}, pages 453--470. Springer, 2024.

\bibitem[Wu and Krahenbuhl(2021)]{wu2021towards}
Chao-Yuan Wu and Philipp Krahenbuhl.
\newblock Towards long-form video understanding.
\newblock In \emph{Proceedings of the IEEE/CVF Conference on Computer Vision and Pattern Recognition}, pages 1884--1894, 2021.

\bibitem[Xu et~al.(2024{\natexlab{a}})Xu, Zhao, Zhou, Lin, Ng, and Feng]{Xu2024PLLaVAP}
Lin Xu, Yilin Zhao, Daquan Zhou, Zhijie Lin, See~Kiong Ng, and Jiashi Feng.
\newblock Pllava : Parameter-free llava extension from images to videos for video dense captioning.
\newblock \emph{ArXiv}, abs/2404.16994, 2024{\natexlab{a}}.

\bibitem[Xu et~al.(2024{\natexlab{b}})Xu, Gao, Gan, Chen, Lai, Gang, Kang, and Dehghan]{xu2024slowfast}
Mingze Xu, Mingfei Gao, Zhe Gan, Hong-You Chen, Zhengfeng Lai, Haiming Gang, Kai Kang, and Afshin Dehghan.
\newblock Slowfast-llava: A strong training-free baseline for video large language models.
\newblock \emph{arXiv preprint arXiv:2407.15841}, 2024{\natexlab{b}}.

\bibitem[Xue et~al.(2022)Xue, Sun, Liu, Fu, Song, Li, and Luo]{Xue2022CLIPViPAP}
Hongwei Xue, Yuchong Sun, Bei Liu, Jianlong Fu, Rui Song, Houqiang Li, and Jiebo Luo.
\newblock Clip-vip: Adapting pre-trained image-text model to video-language alignment.
\newblock In \emph{International Conference on Learning Representations}, 2022.

\bibitem[Yang et~al.(2021)Yang, Miech, Sivic, Laptev, and Schmid]{yang2021just}
Antoine Yang, Antoine Miech, Josef Sivic, Ivan Laptev, and Cordelia Schmid.
\newblock Just ask: Learning to answer questions from millions of narrated videos.
\newblock In \emph{Proceedings of the IEEE/CVF international conference on computer vision}, pages 1686--1697, 2021.

\bibitem[Yang et~al.(2023)Yang, Nagrani, Seo, Miech, Pont-Tuset, Laptev, Sivic, and Schmid]{Yang2023Vid2SeqLP}
Antoine Yang, Arsha Nagrani, Paul~Hongsuck Seo, Antoine Miech, Jordi Pont-Tuset, Ivan Laptev, Josef Sivic, and Cordelia Schmid.
\newblock Vid2seq: Large-scale pretraining of a visual language model for dense video captioning.
\newblock \emph{2023 IEEE/CVF Conference on Computer Vision and Pattern Recognition (CVPR)}, pages 10714--10726, 2023.

\bibitem[Ye et~al.(2024)Ye, Gan, Huang, Ge, Shan, and Tang]{ye2024voco}
Xubing Ye, Yukang Gan, Xiaoke Huang, Yixiao Ge, Ying Shan, and Yansong Tang.
\newblock Voco-llama: Towards vision compression with large language models.
\newblock \emph{arXiv preprint arXiv:2406.12275}, 2024.

\bibitem[Yu et~al.(2022)Yu, Wang, Vasudevan, Yeung, Seyedhosseini, and Wu]{yu2022coca}
Jiahui Yu, Zirui Wang, Vijay Vasudevan, Legg Yeung, Mojtaba Seyedhosseini, and Yonghui Wu.
\newblock Coca: Contrastive captioners are image-text foundation models.
\newblock \emph{Transactions on Machine Learning Research}, 2022.

\bibitem[Zhou et~al.(2021)Zhou, Lin, Li, and Zheng]{zhou2021graph}
Jiaming Zhou, Kun-Yu Lin, Haoxin Li, and Wei-Shi Zheng.
\newblock Graph-based high-order relation modeling for long-term action recognition.
\newblock In \emph{Proceedings of the IEEE/CVF Conference on Computer Vision and Pattern Recognition}, pages 8984--8993, 2021.

\bibitem[Zhu et~al.(2024{\natexlab{a}})Zhu, Lin, Ning, Yan, Cui, HongFa, Pang, Jiang, Zhang, Li, Zhang, Li, Liu, and Yuan]{zhu2024languagebind}
Bin Zhu, Bin Lin, Munan Ning, Yang Yan, Jiaxi Cui, WANG HongFa, Yatian Pang, Wenhao Jiang, Junwu Zhang, Zongwei Li, Cai~Wan Zhang, Zhifeng Li, Wei Liu, and Li Yuan.
\newblock Language{B}ind: Extending video-language pretraining to n-modality by language-based semantic alignment.
\newblock In \emph{The Twelfth International Conference on Learning Representations}, 2024{\natexlab{a}}.

\bibitem[Zhu et~al.(2024{\natexlab{b}})Zhu, Liao, Zhang, Wang, Liu, and Wang]{Zhu2024VisionME}
Lianghui Zhu, Bencheng Liao, Qian Zhang, Xinlong Wang, Wenyu Liu, and Xinggang Wang.
\newblock Vision mamba: Efficient visual representation learning with bidirectional state space model.
\newblock \emph{ArXiv}, abs/2401.09417, 2024{\natexlab{b}}.

\end{thebibliography}
}
\clearpage
\setcounter{page}{1}
\maketitlesupplementary

\section{Implementation Details}
\label{app:implementation_details}

\paragraph{Short-video segmentation.} To optimize training time, we pre-process all data once prior to pre-training. Specifically, for each dataset, the process outlined in \cref{subsec:short_video_representation} is applied only once. This ensures efficient handling of video inputs during the training phase.

\paragraph{Architectural design.} 
We adopt an asymmetric encoder-decoder architecture inspired by~\cite{he2022masked}. The encoder processes only the visible, unmasked tokens from the video input. Positional embeddings are added to each token to encode temporal relationships. The embedded tokens are then passed through a Transformer encoder with $K$ layers, where each layer includes a multi-head self-attention mechanism, a multi-layer perceptron (MLP), and LayerNorm. Unlike ViT architectures designed for fixed-size inputs, such as images, our encoder accommodates varying input lengths, enabling it to handle videos of arbitrary durations. Mask tokens are not used during this stage, ensuring computational efficiency by focusing exclusively on visible tokens.

After the encoder processes the visible tokens, the decoder reconstructs the full sequence by introducing shared, learned mask tokens to replace the missing inputs. These mask tokens are inserted at their respective positions in the sequence. To ensure that the mask tokens carry information about their temporal location, positional embeddings are applied to all tokens in the decoder input, including the mask tokens. Without these positional embeddings, the mask tokens would have no information about their location in the video. The decoder processes the full sequence through a series of Transformer layers. Following~\cite{he2022masked}, it is designed to be shallower than the encoder to maintain computational efficiency while providing accurate reconstructions.

To manage videos of varying lengths, we adapt sequence-processing techniques from BERT~\cite{devlin2019bert}. Each sequence of short-video embeddings is capped at 256 tokens, with shorter sequences padded using a special \texttt{[PAD]} token. During the self-attention process, an attention mask prevents these padding tokens from influencing training. This design ensures that the model maintains computational efficiency while supporting inputs of arbitrary length. Practically, our framework allows for increasing the maximum token limit, enabling the processing of even longer videos. However, since the benchmark datasets used in this study do not exceed 20 minutes, we leave such extensions for future exploration.

Finally, during pre-training, an auxiliary \texttt{[CLS]} token is appended to the input sequence in the encoder. This token serves as a representation of the entire video sequence and is specifically used for downstream tasks, such as classification. In attentive probing fine-tuning, this \texttt{[CLS]} token is adapted to generate task-specific predictions.

\begin{table}[t]
    \caption{\textbf{Pre-training setting.}}
    \vspace{-1.5em}
    \begin{center}
    \small
        \begin{tabular}{l|c}
            \hline
            
            \hline
            
            \hline\\[-3mm]
            \textbf{Config} & \textbf{Value } \\
            \hline
            optimizer & AdamW \\
            base learning rate & 1.5e-4 \\
            weight decay & 0.05 \\
            optimizer momentum & $\beta_1, \beta_2{=}0.9, 0.95$ \\
            batch size & 16 \\
            learning rate schedule & cosine decay \\
            warmup epochs & 40 \\
            epochs  & 150 \\
            number of tokens & 256 \\ 
            short video length & 5 sec \\ 
            \hline 
            
            \hline
            
            \hline
        \end{tabular}

    \end{center}
\vspace{-1.2em}
\label{tab:hyperparams_pretrain}
\vspace{1.5em}
\end{table}

\begin{table}[t]
    \caption{\textbf{Linear probing setting.}}
    \vspace{-1.5em}
    \begin{center}
    \small
        \begin{tabular}{l|c}
            \hline
            
            \hline
            
            \hline\\[-3mm]
            \textbf{Config} & \textbf{Value } \\
            \hline
            optimizer & Adam \\
            base learning rate & 1e-4 \\
            
            optimizer momentum & $\beta_1, \beta_2{=}0.9, 0.999$ \\
            batch size & 16 \\
            epochs  & 30 \\
            \hline 
            
            \hline
            
            \hline
        \end{tabular}
    \end{center}
\vspace{-1.2em}
\label{tab:hyperparams_lp}
\vspace{1.5em}
\end{table}

\begin{table}[t]
    \caption{\textbf{Attentive probing setting.}}
    \vspace{-1.5em}
    \begin{center}
    \small
        \begin{tabular}{l|c}
            \hline
            
            \hline
            
            \hline\\[-3mm]
            \textbf{Config} & \textbf{Value } \\
            \hline
            optimizer & Adam \\
            learning rate & 1e-3 \\
            learning rate schedule & ExponentialLR, $\gamma{=}0.9$ \\
            optimizer momentum & $\beta_1, \beta_2{=}0.9, 0.999$ \\
            batch size & 16 \\
            epochs  & 20 \\
            \hline 
            
            \hline
            
            \hline
        \end{tabular}
    \end{center}
\vspace{-1.2em}
\label{tab:hyperparams_ap}
\vspace{-0.5em}
\end{table}

\begin{table*}[t]
    \caption{\textbf{Additional LVU benchmark results.} In the main paper, we provided the average Top-1 accuracy results obtained by LV-MAE with linear probing (LP) and attentive probing (AP) with random masking. Here, we provide extended results per task. Masking indicates the masking technique that was applied. ``Dir.'' and ``Rel.'' refer to ``Director'' and ``Relationship,'' respectively.}
    \label{tab:lvu_benchmark_app}
    \centering
    \small
    \setlength{\tabcolsep}{6.5pt}
    \begin{tabular}[t]{l|c|cccc|ccc|c}
            \hline 
            
            \hline
            
            \hline
        \multirow{2}*{\textbf{Method}} & \multirow{2}*{\textbf{Masking}} & \multicolumn{4}{c|}{\textbf{Metadata $\uparrow$}} & \multicolumn{3}{c|}{\textbf{Content $\uparrow$}} & \multirow{2}*{\textbf{Avg.}} \\
        & & \textbf{Dir.} & \textbf{Genre} & \textbf{Writer} & \textbf{Year} & \textbf{Scene} & \textbf{Speak} & \textbf{Rel.} & \\
        \hline
        $\text{LV-MAE}_{\text{InternVideo2}}$(LP)  & Semantic & 55.14 &  58.56 & 49.40 & 43.26 & 72.84 & 37.77 & 51.22 & 52.60 \\
        $\text{LV-MAE}_{\text{InternVideo2}}$(LP)  & Random & 57.01 &  57.36 & 52.38 & 49.65 & 70.37 & 39.89 & 48.78 & 53.63 \\
        $\text{LV-MAE}_{\text{InternVideo2}}$(AP) & Random & 54.21 &  66.95 & 55.36 & 56.03 & 80.25 & 42.55 & 53.66 & 58.43 \\
        $\text{LV-MAE}_{\text{LanguageBind}}$(LP)  & Semantic & 71.03 & 67.47 & 54.76 & 53.90 & 67.90 & 42.02 & 56.09 & 59.02 \\
        $\text{LV-MAE}_{\text{LanguageBind}}$(LP)  & Random & 71.96 & 67.12 & 55.36 & 53.19 & 71.60 & 37.76 & 51.22 & 58.32 \\
        $\text{LV-MAE}_{\text{LanguageBind}}$(AP)  & Random & 78.50 & 69.17 & 60.12 & 61.70 & 72.84 & 39.36 & 56.09 & 62.54 \\
            \hline 
            
            \hline
            
            \hline
    \end{tabular}
\end{table*}

\paragraph{Pre-training hyperparameters.} 
Our pre-training hyperparameters are detailed in Table~\ref{tab:hyperparams_pretrain}. We closely follow the hyperparameter settings from~\cite{he2022masked} with a few adjustments. Specifically, we use a smaller batch size and reduce the number of epochs during pre-training. We set the limit for the number of tokens to 256. Finally, we set the short video segment length to five seconds.
The choice of a five-second segment length balances two considerations: it is short enough to capture low-level spatiotemporal patterns effectively using an off-the-shelf frozen model and sufficiently long to reduce the number of tokens required for longer videos. In future work, we plan to analyze the impact of segment lengths (e.g., 10–15 seconds) on performance, as longer segments could offer greater efficiency for processing extended videos. However, this may risk reduced performance from the frozen model due to challenges in extracting representations from longer, more complex segments.

\section{Training on Downstream Tasks}
\label{app:results_downstream_tasks}

We experiment with two approaches to train our frozen model to solve downstream tasks utilizing its latent representations.

\paragraph{Linear probing.} For linear probing, we append a simple linear layer to the encoder. This layer operates on the global average pooling of the latent representations $\mathcal{Z}$. The linear layer is optimized with cross-entropy loss. We report the optimizer and other hyperparameters we use in  Table~\ref{tab:hyperparams_lp}.

\paragraph{Attentive probing.} To implement attentive probing~\cite{yu2022coca, lee2019set}, we add a lightweight transformer encoder layer to the pre-trained model. This additional block consists of a multi-head self-attention mechanism, an MLP, and LayerNorm. The layer is fine-tuned exclusively on task-specific datasets, focusing on adapting the \texttt{[CLS]} token to generate final predictions through a linear classifier optimized with cross-entropy loss. This approach minimizes computational overhead while effectively tailoring the model for specific tasks. We report the optimizer and other hyperparameters we use in Table~\ref{tab:hyperparams_ap}.

\section{Extended LVU ablation}
\label{app:extended_lvu_ab}

In Table~\ref{tab:lvu_benchmark_app}, we provide full results of the LVU benchmark for the reported average classification score from Table~\ref {tab:masking_type}.

\paragraph{Impact of Segment Length.}
In the main paper, we report results obtained by partitioning each long
video into \emph{five} second clips.  
To assess the sensitivity of our method to this design choice, we also
trained models on longer fixed-length clips (10 seconds and 15 seconds) as well as
on variable-length, shot-based segments.  
Table~\ref{tab:segment_length} summarizes the average performance across all
LVU downstream tasks. The \emph{five} second configuration remains the most
effective, achieving an average score of 63.40 and consistently
outperforming the longer and shot-based alternatives.
Finally, we examined the use of \emph{overlapping} \emph{five} second clips; this introduces redundant context and reduces the average score by
4.5, from 63.40 to 58.90.

\begin{table}[h]
\centering
\begin{center}
\caption{\textbf{Impact of Segment Length.} The average Top-1 accuracy results obtained by LV-MAE on the LVU benchmark using different segment lengths.}
\begin{tabular}{cccc}
\hline

\hline

\hline
\textbf{5 seconds} & \textbf{10 seconds} & \textbf{15 seconds} & \textbf{shots}\\
\hline
63.40 & 61.43 & 61.05 & 62.17 \\
\hline

\hline

\hline
\label{tab:segment_length}
\end{tabular}
\end{center}
\end{table}

\paragraph{Architecture and Clip-Length Ablations.} Our performance gains stem from both the two-stage architecture and long-video training capability. While prior works are limited to $\thicksim$60 frames, our method can process much longer sequences. To highlight the importance of this capability, we cap the training
clips at \emph{five minutes}, which lowers the average LVU score to
55.58\,\%. Comprehensive ablations, removing the MAE stage or replacing it with plain Transformer or Mamba backbones, are reported in
Table~\ref{tab:arch_lvu_abl}. The results demonstrate that
\emph{both} the MAE formulation and exposure to extended temporal context are critical for achieving state-of-the-art performance.

\begin{table*}[h]
    \centering
    \caption{Results on the LVU benchmark using different architectures and clip-length ablations.}
    \label{tab:arch_lvu_abl}
    \begin{tabular}[t]{l|c|cccc|ccc|c}
            \hline 
            
            \hline
            
            \hline
        \multirow{2}*{\textbf{Method}} & \multirow{2}*{\shortstack{\textbf{FB}}} & \multicolumn{4}{c|}{\textbf{Metadata $\uparrow$}} & \multicolumn{3}{c|}{\textbf{Content $\uparrow$}} & \multirow{2}*{\textbf{Avg.}} \\
        & & \textbf{Dir.} & \textbf{Genre} & \textbf{Writer} & \textbf{Year} & \textbf{Scene} & \textbf{Speak} & \textbf{Rel.}\\
        \hline
        ViS4mer & \xmark & 62.61 & 54.71 & 48.80 & 44.75 & 67.44 & \underline{40.79} & 57.14 & 53.7 \\
        VideoMamba & \xmark & \underline{67.29} & 65.24 & \underline{52.98} & 48.23 & 70.37 & 40.43 & \textbf{62.50} & \underline{58.1} \\
        \hline
        LanguageBind-Transformer & \xmark & 24.30 & 57.88 & 16.07 & 18.44 & 35.80 & 32.45 & 51.22 & 33.7 \\
        LanguageBind-Mamba & \xmark & 61.68 & \underline{70.38} & 51.78 & \underline{56.74} & \textbf{74.04} & 38.83 & 43.90 & 56.8  \\
        $\text{LV-MAE} ({\text{frame-MAE feature extractor)}}$ & \cmark & 64.49 & 62.50 & 49.4 & 48.23 & 67.90 & 36.70 & 51.22 & 54.3  \\
        \hline
        \rowcolor{Gray!15} $\text{LV-MAE}_{\text{LanguageBind}}$(Ours) & \cmark & \textbf{77.57} & \textbf{71.57} & \textbf{64.28} & \textbf{58.15} & \underline{72.84} & \textbf{40.95} & \underline{58.53} & \textbf{63.4}  \\
            \hline 
            
            \hline
            
            \hline
    \end{tabular}
\end{table*}

\begin{figure}[h]
  \centering
  \includegraphics[width=1.\linewidth]{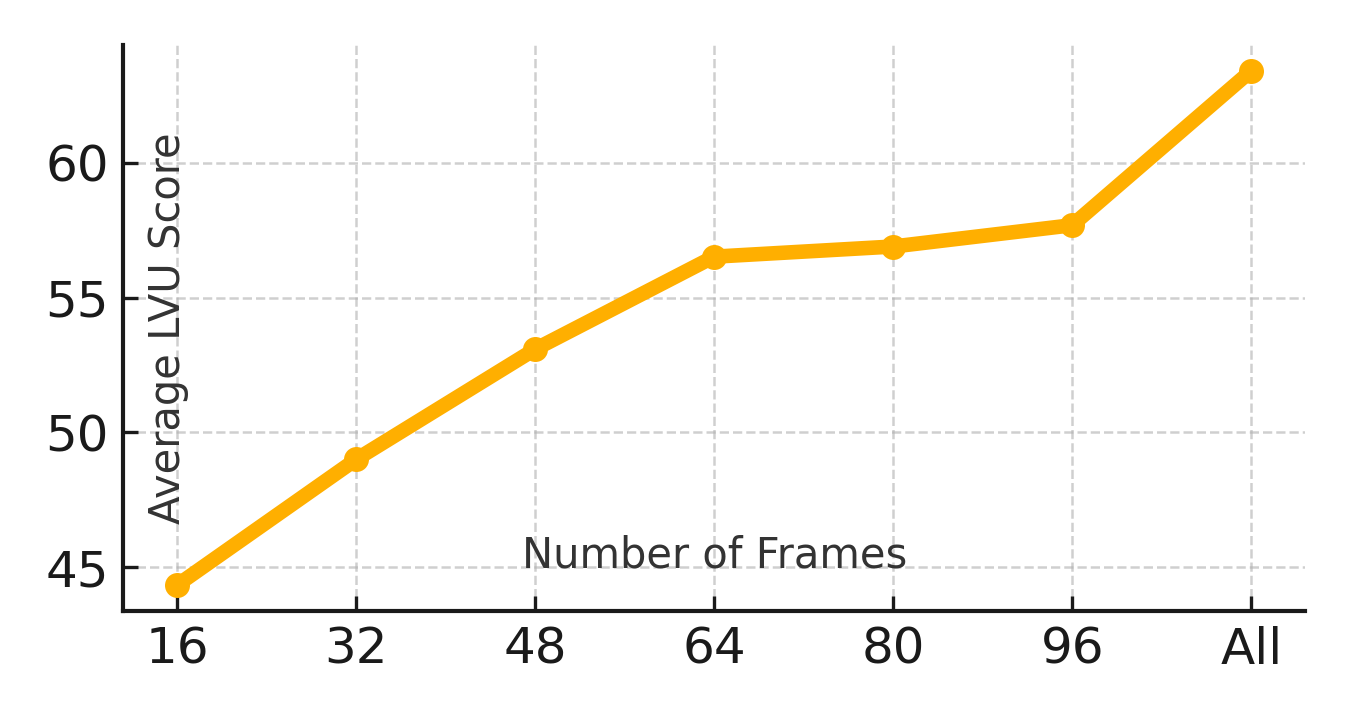}
  \caption{Average performance of LVU benchmark rises monotonically as the frame count increases.}
  \label{fig:lvu_func_len}
\end{figure}

\paragraph{Impact of Input Frame Count.}
To quantify the role of temporal coverage, we trained models with
varying numbers of input frames; the resulting trend is depicted in
Fig.~\ref{fig:lvu_func_len}.  Performance rises monotonically as the frame
count increases, confirming that a broader temporal window enables the
model to capture motion cues more effectively.  We extract 5-second clip embeddings with clip count varying by video length. All methods use identical input resolution, but ours processes significantly more frames than prior work ($\thicksim$60 frames limit) efficiently, which is a key contribution enabling better temporal understanding.

\section{Short-Video Performance} 
\label{app:short_video_perf}

Our approach preserves performance on short-video understanding tasks. Specifically, we preserve the accuracy of LanguageBind on Kinetics-400 (600)~\cite{Kay2017TheKH}.
We achieved 78.2\% on K400 and 79.1\% on K600 compared to 77.6\% and 79.5\% in LanguageBind.

\section{Additional Interpretable Predictions Results}
\label{app:interpretable_pred}

\begin{figure*}[ht]
    \centering
    \includegraphics[width=1.\textwidth]{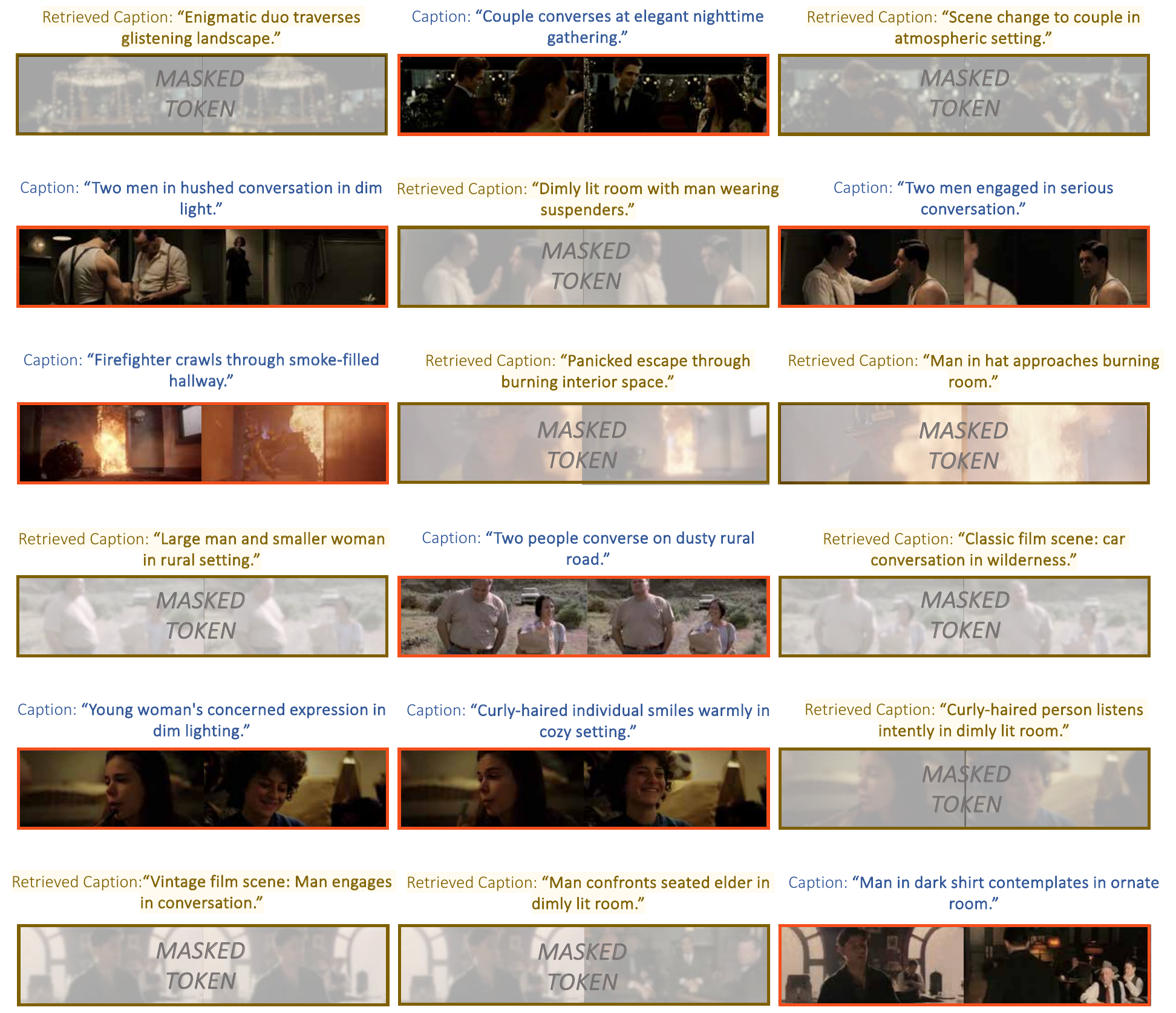}
    \caption{\textbf{Interpretable predictions -- additional examples}: Each row visualizes three consecutive five-second segments. Above each segment, we show the original caption for the visible tokens and the retrieved caption for the reconstructed masked tokens. As shown, the model successfully reconstructs the semantic meaning of the masked embeddings, offering insight into the model's effectiveness and capabilities.}
    \label{fig:interp_res_app}
\end{figure*}

In Fig.~\ref{fig:interp_res_app}, we attach additional examples of the interpretable predictions experiment from \cref{sec:exp_interpretabitlity}.

\section{Additional Related Work}
\label{app:related}

\paragraph{Short-video understanding methods.} Numerous models have been proposed for short-video understanding, achieving remarkable performance on tasks such as action recognition \cite{arnab2021vivit}, video classification \cite{fan2021multiscale}, text-video retrieval \cite{luo2021clip4clip}, video captioning \cite{yang2021just}, and video question answering \cite{lei2021less}. Multimodal models like CLIP \cite{radford2021learning}, InternVideo \cite{Wang2024InternVideo2SV}, and LanguageBind \cite{zhu2024languagebind} have demonstrated strong capabilities in aligning video and language representations. In this work, we leverage the output embedding of these models to learn long-video representations. 

\paragraph{Long-video language understanding methods.} Several recent works have advanced the field of video understanding using large language models. Video-XL~\cite{shu2024video} and LongVLM~\cite{weng2024longvlm} both tackle the challenge of processing long-form videos, with Video-XL focusing on hour-scale video understanding and LongVLM proposing efficient mechanisms for extended video content. LongVILA~\cite{chen2024longvila} further contributes to this direction by scaling visual language models for long video comprehension. In the domain of efficient video processing, VoCo-LLaMA~\cite{ye2024voco} explores video compression using large language models, while LLaMA-VID~\cite{li2024llama} proposes a compact two-token representation for video content. SlowFast-LLaVA~\cite{xu2024slowfast} provides a training-free baseline approach for video large language models. Addressing temporal aspects, TimeChat~\cite{ren2024timechat} develops time-sensitive capabilities for video understanding, while LITA~\cite{huang2024lita} focuses on precise temporal localization within videos. In contrast to these works that focus on video-language integration, we explore long-video masked-embedding autoencoders in long-form video classification tasks and aim to find effective representation learning methods specifically designed for long-form videos.







\end{document}